\documentclass[bst-sn-basic]{sn-jnl}
\pgfplotsset{compat=1.17}
\usepackage{subfigure}
\usepackage{makecell}
\usepackage[sort&compress,numbers]{natbib}
\usepackage{soul,color}
\usepackage{lineno}

\def\eg{\emph{e.g}} 
\def\ie{\emph{i.e}}

\jyear{2021}%

\begin{document}

\title{Rapid Person Re-Identification via Sub-space Consistency Regularization}

\author*[1]{\fnm{Qingze} \sur{Yin}}\email{qingzeyin@njust.edu.cn}

\author[2]{\fnm{Guan'an} \sur{Wang}}\email{guan.wang0706@gmail.com}

\author[3]{\fnm{Guodong} \sur{Ding}}\email{dinggd@comp.nus.edu.sg}

\author[4]{\fnm{Qilei} \sur{Li}}\email{q.li@qmul.ac.uk}

\author[4]{\fnm{Shaogang} \sur{Gong}}\email{s.gong@qmul.ac.uk}

\author[1]{\fnm{Zhenmin} \sur{Tang}}\email{Tzm.cs@njust.edu.cn}

\affil*[1]{\orgdiv{the Collaborative Innovation Center of Social Safety Science and Technology}, \orgname{Nanjing University of Science and Technology}, \orgaddress{\street{Xiaolingwei Street}, \city{Nanjing}, \postcode{210094}, \state{Jiangsu}, \country{China}}}

\affil[2]{\orgdiv{Institute of Automation}, \orgname{Chinese Academy of Sciences}, \orgaddress{\city{Beijing}, \postcode{100045}, \state{State}, \country{China}}}

\affil[3]{\orgdiv{School of Computing}, \orgname{National University of Singapore}, \country{Singapore}}

\affil[4]{\orgdiv{Department}, \orgname{the Queen Mary University of London}, \orgaddress{\street{Street}, \city{London}, \postcode{E1 4NS}, \state{London}, \country{UK}}}


\abstract{
Person Re-Identification (ReID) matches pedestrian across disjoint cameras.
Existing ReID methods adopting real-value feature descriptors have achieved high accuracy, but they are low in efficiency due to the 
slow Euclidean distance computation as well as complex quick-sort algorithms.
Recently, some works propose to yield binary encoded person descriptors which instead only require  
fast Hamming distance computation and simple counting-sort algorithms.
However, the performances of such binary encoded descriptors, especially with short code (\eg, 32 and 64 bits), are hardly satisfactory given the sparse binary space.
To strike a balance between the model accuracy and efficiency, we propose a novel Sub-space Consistency Regularization (SCR) algorithm that can speed up the ReID procedure by $0.25$ times than real-value features under same dimensions whilst maintain a competitive accuracy, especially under short codes.
SCR transforms real-value features vector (\eg, 2048 float32) with short binary codes (\eg, 64 bits) by first dividing real-value features vector into $M$ sub-spaces, each with $C$ clustered centroids. Thus the distance between two samples can be expressed as the summation of respective distance to the centroids, which can be sped up by offline calculation and maintained via a look-up-table.
On the other side, these real-value centroids help to achieve significantly higher accuracy than using binary code.
Lastly, we convert the distance look-up-table to be integer and apply the counting-sort algorithm to speed up the ranking stage. 

We also propose a novel consistency regularization with an iterative framework. 
Experimental results on Market-1501 and DukeMTMC-reID show promising and exciting results. Under short code, our proposed SCR enjoys Real-value-level accuracy and Hashing-level speed.
}


\maketitle

\section{Introduction}
\label{sec:intro}

Person Re-Identification (ReID)~\cite{bookperson,zheng2016person} is a branch of computer vision task that explicitly deals with person matching cross non-overlapping camera installations. Recent ReID models have made a impressive progress due to the rapid development of deep learning~\cite{wang2019aligngan,wang2020crossmodality,Zhou2021,Yin2021,GraphSampling,LI202146,9495801,9463771,Li2021,yao2021non,yao2021jo}.
However, the critical issue of person ReID in the real world is far from being resolved. In addition to high performance, quick retrieving speed requires improvement when applying ReID in an open-world setting.

Most existing ReID approaches~\cite{Wang2021,Zhao2019,Xiang2020,Feng2019,ding2019feature,wu2022camera,wang2020jislpp,math10101654} achieve a promising accuracy performance using Euclidean distance to measure the similarities of high-dimensional real-value person descriptors.
Analogous to image retrieval,  finding the matched image of a given query in a large gallery first involves calculating pairwise distances in the feature space and then sort per query to obtain a ranking list. 
Measuring the Euclidean distance between two high-dimensional descriptors is computational expensive. Despite that quick-sort algorithm
~\cite{hoare1962quicksort} has been adopted to boost the ranking process, it still has the time complexity of $\mathcal{O}(N\log N)$, which would also become costly when the gallery size $N$ is large. 
The detailed speed comparisons can be viewed in Table~\ref{tab:rankingtime}.

\begin{table}[h]
\centering
\caption{Comparisons for ranking speed per query image}
\label{tab:rankingtime}

\begin{tabular}{@{}lcc@{}}
\toprule
\multirow{2}{*}[-4pt]{Gallery size}   & \multicolumn{2}{c}{Query Time (s)} \\ \cmidrule(l){2-3} 
                    &  Quick-Sort        & Counting-Sort          \\ \midrule
$1\times 10^{3}$ & $3.4\times 10^{-3}$ & $4.7\times 10^{-4}$ \\ 
$1\times 10^{4}$ & $1.0\times 10^{-1}$ & $2.7\times 10^{-3}$ \\ 
$1\times 10^{5}$ & $4.3\times 10^{-1}$ & $2.7\times 10^{-2}$ \\  
$1\times 10^{6}$ & $6.4\times 10^{0}$ & $2.6\times 10^{-1}$ \\ 
$1\times 10^{7}$ & $1.1\times 10^{2}$ & $2.7\times 10^{0}$ \\ \midrule
\multirow{2}{*}{\begin{tabular}[c]{@{}l@{}}Per Sample\\ Complexity\end{tabular}} &     -      &     $2.6\times 10^{-7}$        \\
                    &      $O\left ( NlogN \right )$     &   $O\left ( N \right )$        \\ \bottomrule
\end{tabular}

\end{table}

This has encouraged the ReID community to pay attention to fast ReID~\cite{wang2020faster,7532465}, seeking to speed up the ReID process while sustaining a high accuracy. 
In practice, most existing fast ReID methods~\cite{8100049,7532465,fwPerceptual2017,8784980,Wu2018StructuredDH,Zheng2016LearningCB,fz2017part,wang2021abml} adopt hashing algorithms to generate compact and efficient binary codes as the image descriptor to replace the real-valued features. Binary codes can represent long vectors with fewer bits and achieve higher similarity comparison speed based on Hamming distance calculation. 
Counting-sort, compared to quick-sort, has less time complexity as it creates a number array and counts to sort. It owns a linear complexity concerning the gallery size $\mathcal{O}(N)$.
On another note, counting-sort can only take \textsl{int} type of vectors as input, making it suitable for sorting binary codes. In fast ReID works, Hamming distance and counting-sort~\cite{Bajpai2014ImplementingAA} algorithm are used to obtain the final ranking list.

Even though binary codes can significantly speed up the process, the direct conversion of extracted real-value features to binary code features would lead to a significant ReID performance drop, especially when the coding bits are small, \eg, 32 and 64, as have been reported in previous works~\cite{wang2020faster}. 
Wang \emph{et al.} proposes a CtF~\cite{wang2020faster} method, which using self-distillation learning to learn a shorter code mimicked from a longer code. The Rank-1 and mAP accuracies of CtF adopting with 2048-bits are $1.56\times$ and $2.26\times$ higher than CtF adopting with 32-bits, which demonstrates that the shorter code length generates worse performance. 
ReID in its essence is a fine-grained retrieval task which sets high requirements for the model to disambiguate between person images with large intra-class variations and low inter-class variations. In this regard, it is expected for binary codes
to yield inferior performances as they inherently have much weaker representation capability compared to real value features, especially when the code bits are small.
Above observations have motivated us to focus our research attention on improving the performance for person descriptors, especially when bits are low, and at the same time achieve comparable speed with binary code.

The speed of ReID relies on two main aspects: pairwise distance measurement and ranking speed.
In order to increase the computation speed, we propose to reduce the complexity of the similarity measurement by introducing sub-spaces. The prevailing similarity measurement is done via the Euclidean distance which is calculated in a vector-to-vector fashion and we denote this as instance-based. The complexity would be huge when the image numbers are large. In this regard, we propose to quantize the original image feature vector into multiple sub-spaces and find cluster centroid in each sub-space which serves as the basis for similarity calculation. More concretely, given a set of sub-space centroids, we first find for image pairs their respective sub-space centroids, then regard the sum of distances between their centroids as their distance. The centroid based formulation greatly reduces the calculation as the number of centroids are way less than the demension of the feature vector.
On the other side, to improve the ranking speed, we resort to build a look-up-table (LUT) to memorize the pre-calculated query-gallery pair-wise distances which allows for faster searching speed in the distance matrix based on their respective centroid index. By extension, we can quantize the distance values into \textsl{int} type and apply counting-sort algorithm to further decrease the ranking cost.

The above mentioned methods aim at improving computation speed and seeking better ranking strategies. Another goal of a fast person ReID is to sustain high performance. 
Instance-based distance~\cite{Luo_2019_CVPR_Workshops} achieves high performance as it is computed directly based on the real-value feature vectors between the query and gallery set. As the above analysis introduced, measurement by using Euclidean distance~\cite{Ratcliffe1994FoundationsOH} on real-value features leads to a complex computing process. To increase speed, we choose centroids-based distance~\cite{lin2019bottom} instead of instance-based distance. While our effort for fast ReID imposes centroid-based distance, which bypasses the element-wise calculation with the distance between centroids. It is less accurate and would lead to a performance drop. Similarity structure in the Euclidean space preserves a more rigorous form of sample relations. Thus, we plan to bring closer two types of distance matrices, \ie, instance-based and centroid-based, to inject more precision into the centroid-based matrix. The coarser centroid-based similarity matrix can thus retain a high-performance value during testing. Specifically, we propose a consistency regularization loss forcing the similarity outputs from two different measurements to be consistent. The proposed consistency regularization loss has made the centroid-based distance closer to the instance-based distance during the training iterations.

Combining the two keys aspects, we propose in this paper a Sub-space Consistency Regularization (SCR) model to perform fast ReID. On the one hand, our proposed SCR model first divides the extracted feature vectors into multiple sub-spaces and creates a distance look-up-table for memorizing the similarity calculation based on the sub-space centroids to speed up the testing speed. On the other hand, we propose a consistency regularization loss to guide the model learning towards consistent similarity structures for both instance-based and centroid-based measurements to retain an accuracy even with low bits of encoding.  

We summarize our contributions as follows:
\textbf{(1)} We propose a novel fast ReID model named Sub-space Consistency Regularization (SCR) which targets to speed up ReID and at the mean time retain a high accuracy, especially with low bit binary codes.
\textbf{(2)} We propose to divide person descriptors into multiple sub-spaces where a look-up-table based on sub-space centroids distances can be used for fast construction of sample distances. We further show a faster ranking speed can be achieved with our quantized version of look-up-table combined with counting-sort algorithm.
\textbf{(3)} We propose a sub-space consistency regularization loss to transfer the rigorous Euclidean distance information to the quick centroid-based distance calculation. 
\textbf{(4)} Extensive experiments on two ReID benchmarks Market-1501~\cite{7410490} and DukeMTMC-reID~\cite{zheng2017unlabeled} show that our model significantly outperforms the state-of-the-arts in terms of Rank-1 score and speed time.

The remaining contents of this paper are organized as follows: Sec.~\ref{sec:background} reviews the related work in ReID. Sec.~\ref{sec:method} elaborates the proposed framework and algorithm. Experiments and discussions are presented in Sec.~\ref{sec:exp}. Finally, Sec.~\ref{sec:con} concludes this work.

\section{Related Works}
\label{sec:background}

ReID is a person retrieval task which aims to match the same pedestrian images across disjoint cameras from a large dataset. Therefore, the high accuracy and better efficiency for matching are the two main goals in ReID task. In this regard, we review the related works as accuracy oriented and efficiency oriented.

\textbf{Accuracy oriented ReID} The main challenge on the ReID accuracy comes from the different camera viewpoints, illuminations, low resolution, poses variations and occlusions. These factors cause a large intra-class gap and a small inter-class gap. There are three main streams of methods: hand-crafted, deep learning, and metric learning. Feature learning approaches seek to learn better feature representations. Hand-crafted methods~\cite{7298832,5539926} aim to produce a robust feature representation. Such as, SDALF method~\cite{7298832} proposes a novel feature extraction and matching strategy method based on the localization of perceptual concerned person parts to achieve robustness of illumination variations, viewpoint and pose.
While recent deep methods~\cite{9566578,wang2020cvpr,Luo_2019_CVPR_Workshops,qian2018pose,pcb2018,9463771,LI202146,9495801} learn robust and discriminative feature representations with deep neural networks. Sun~\emph{et al.}~\cite{pcb2018} proposes a novel approach named PCB, which improves the part-based feature learning by proposing a refined part pooling method. Qian~\emph{et al.}~\cite{qian2018pose} proposes an adversarial image generation model for pose normalization. The generated images are identity-preserving, realistic, and pose controllable. Luo~\emph{et al.}~\cite{Luo_2019_CVPR_Workshops} collects and combines many state-of-the-art person ReID training tricks as a strong person ReID baseline, which achieves 94.5\% rank-1 and 85.9\% mAP on Market-1501 with global features. Yin~\emph{et al.}~\cite{9463771} proposes a novel network that uses multi-view labels voting to assign soft pseudo labels and performs a confidence-based clustering. Li~\emph{et al.}~\cite{LI202146} proposes an end-to-end network that learns with cross-adversarial consistency (CAC) and consistency self-prediction (CSP) constraint. CAC improves the domain invariance of the learned features, while CSP increases the feature discrimination. Li~\emph{et al.}~\cite{9495801} proposes a novel triple adversarial learning and a multi-view imaginative reasoning network (TAL-MIRN). The proposed network comprises two modules to extract comprehensive descriptions of pedestrian appearance and alleviate the problematic domain shift.
Metric learning algorithms~\cite{ding2019a,6247939,7410777,6226421,7780518}
leverage the similarities between sample pairs~\cite{6226421} 
or triplets~\cite{7780518} to pull closer person images with the same identity and push further away images of different identities.
Those methods can obtain a high accuracy based on their real-value person descriptors, however, they are time-consuming during testing due to the slow speed of similarity calculation as well as ranking.

\textbf{Efficiency oriented ReID} Existing approaches addressing the efficiency problem mainly applies hashing algorithms that can achieve a fast speed for the retrieval process and meanwhile obtain a reasonable accuracy. Specifically, in replacement of inefficient real-value based Euclidean distance and quick-sort algorithm, the binary-code based methods adopt hamming distance and counting-sort that are much more efficient. For example, 
TDDH~\cite{8876675} unifies the extracted discriminative feature representation and the binary codes which are transformed with the independence and balance properties in an end-to-end pipeline. CSBT~\cite{8100049} applies sub-space projection to alleviate the cross-view variations by maximizing intra-class compactness and inter-class well-separation. ABC~\cite{8784980} distinguishes the real-value feature from binary codes by training a discriminator network. CtF~\cite{wang2020faster} utilizes shorter codes to rank coarsely and uses longer codes to refine the top candidates, where the shorter codes are encouraged to mimic the longer codes by self-distillation learning.
These methods with binary codes are quick in searching, but would lead to an unsatisfying performance when the codes are short. In this work, we bridge the distances gap between centroid-to-centroid and vector-to-vector by learning joint feature spaces with our proposed sub-space consistency regularization loss.

\section{The Proposed Method}
\label{sec:method}
\begin{figure*}[t]
\centerline{\includegraphics[height=7.5cm,width=10cm]{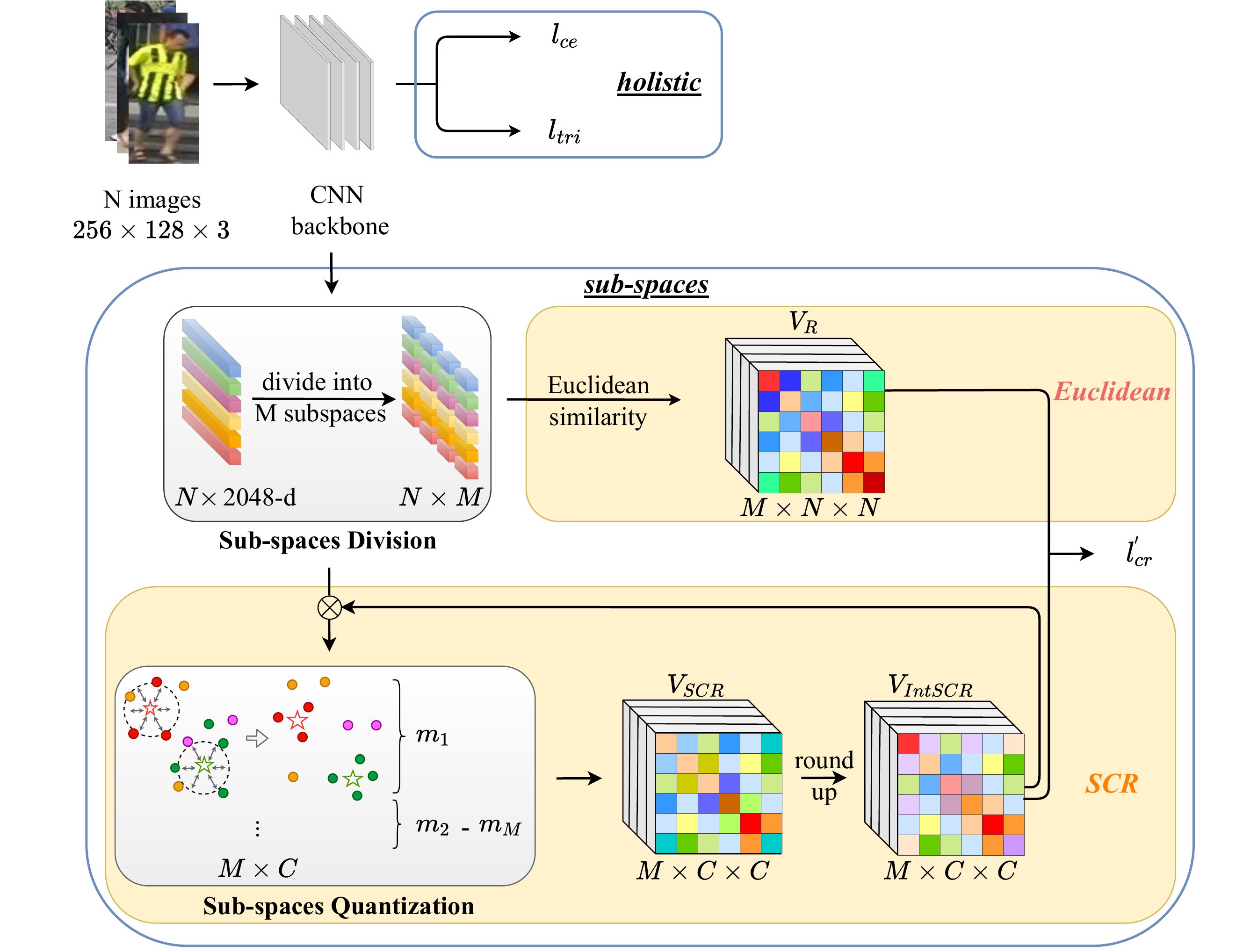}}

\caption{SCR framework. Image features are obtained via CNN backbone and then fed into two branches: global branch (holistic) for calculating ReID cross-entropy loss and triplet loss, local branch (sub-spaces) for consistency regularization learning. }
\label{fig:pq-reid}
\end{figure*}
The proposed Sub-space Consistency Regularization (SCR) model comprises three main parts, \ie, the backbone, holistic branch and sub-spaces branch. Given a selected backbone network, the holistic branch is introduced to learn robust and discriminative person descriptors; while the sub-spaces branch is imposed to perform the fast inference of the ReID task and hence the focus of this paper. 
The sub-spaces branch also includes two sub-branches: Euclidean and SCR branches shown in yellow blocks separately. As shown in the sub-spaces branch in Fig.~\ref{fig:pq-reid}, the input features vectors in dimension of $N\times 2048$ are divided into $M$ sub-spaces which is $N\times M$.  
Fig.~\ref{fig:pq-reid} depicts the overall framework. 
In the sub-space branch, there are two sub-branches for generating $V_{IntSCR}$ and $V_{R}$. In the SCR branch, an exemplar (denoted by $\star$) shows the quantization process per sub-space, and $\bigotimes$ represents the interval center re-initialization. In sub-space quantization, $m_{m}$ represents the $m$-th sub-space and each sub-space includes 256 centers.

\subsection{Backbone Model and Holistic Branch} 
\label{subsec:cnn}
Given a training set $X=\left \{ x_{1} ,x_{1} ,...,x_{N} \right \}$ of $N$ total images from $Y$ identities, a pre-trained ResNet50~\cite{7780459} is adopted as the CNN backbone which extracts a 2048-dimensional feature for each image.
Following previous work for person ReID~\cite{Tan2020MHSANetMS,liiccv2019}, 
in holistic branch, we apply the commonly adopted cross-entropy and triplet loss together to enforce the  feature learning written as: 
\begin{equation}
    \mathcal{L} = l_{ce} + l_{tri}
\end{equation}

More concretely, we write the cross-entropy loss 
\begin{equation}
    l_{ce}=\frac{1}{N}\sum_{i=1}^N p(y_i)\log(f(\phi( x_{i};\theta);\omega)),
\end{equation}
where $\theta$ denotes the parameter for backbone model $\phi(\cdot)$ and $f(\cdot;\omega)$ is the last fully-connected layer parametrized by $\omega$.
A triplet loss $l_{tri}$ is imposed on the derived feature vector $v_{x}$ generated from the base CNN model in holistic branch. $l_{tri}$ would minimize the intra-class distinctness while maximizing the inter-class discrepancy. For each image $x$, we sample a negative image $x_{neg}$ which has different identity label and a positive image $x_{pos}$ which has the same identity label to form a triplet tuple. Then the distances between $x$ and $x_{neg}$(or $x_{pos}$) can be calculated as:

\begin{equation}
    d_{x,neg} =\left \| v_{x}-v_{x,neg} \right  \|_{2},  \; d_{x,pos} =\left \| v_{x}-v_{x,pos} \right \|_{2},
\end{equation}
Hence, the triplet loss $l_{tri}$ is formulated as:
\begin{equation}
    l_{tri}=\max ( 0,m+d_{x,pos}-d_{neg} ),
    \label{tri}
\end{equation}
where $m\in \mathbb{R}>0$ and it is a margin parameter which can enforce the separation between negative and positive image pairs.
With the above loss formulation, we can learn the discriminative feature representations which will be used in our sub-spaces branch. 

\subsection{Sub-space Consistency Regularization}
\label{subsec:defin}
The Sub-spaces branch builds two heterogeneous distance matrixes, \ie, Euclidean-based and Quantization-based, and the sub-space consistency regularization loss is imposed to align above two distance matrixes such that a high accuracy can be obtained.
Given image features extracted by the backbone network, we first split each feature vector uniformly to form a fixed number of $M$ sub-spaces, \ie, each sub-space is spawned by sub-portions of original feature vector. Within each sub-space, we create two types of pairwise distance matrix between each batch of person images.

\textbf{Euclidean-based distance matrix $V_R$}.
The similarity of image pair $d\left ( x_{i},x_{j} \right )$ is equal to the summation of each corresponding sub-space similarity $\sum_{m=1}^{M}d\left ( x_{i}^{m},x_{j}^{m} \right )$. Where we choose to use the Euclidean distance to calculate and store the sub-space similarities in the matrix $V_{R}$. Therefore the size of this look-up-table $V_{R}$ is $M \times N \times N$.

\textbf{Quantization-based distance matrix $V_{SCR}$}.
For the quantization distance matrix, within each sub-space, $k$-means clustering algorithm~\cite{1641018} is applied to find in total $C$ centroids. 
Then a vector can be represented by a short code which consists of its separated sub-space quantization indices. 

\begin{equation}
Vector_{x_{i}}=Combine\left ( M_{x_{i}^{1}}^{1}, M_{x_{i}^{2}}^{2},..., M_{x_{i}^{m}}^{m}  \right ), M_{x_{i}^{1}}^{1}=C_{x_{i}^{1}}^{1}
\label{eq:quantization}
\end{equation}

\begin{figure*}[ht]
\centering
\includegraphics[height=6cm,width=9cm]{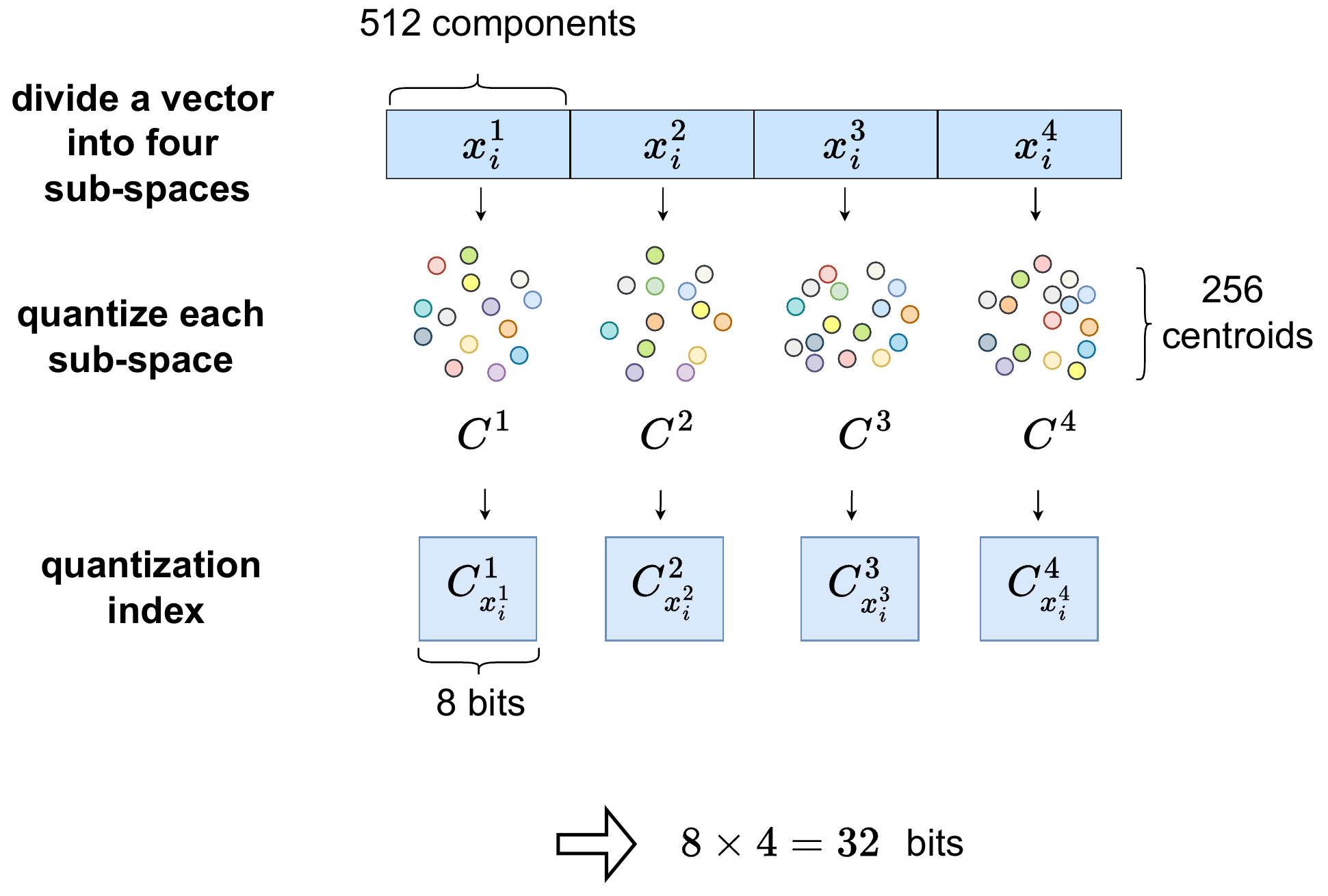}
\caption{Quantization in SCR.}
\label{fig:prin}
\end{figure*}

Where $C$ is consists of 256 centeroids. The quantization indices are defined as their corresponding clustering centroids indices. For example, $C_{x_{i}^{1}}^{1}$ represents the quantization indices of the vector's first sub-space. Thus, the original distance can be replaced with the summation of each sub-space's centroids pre-computed Euclidean distances. The computation of distances based on their short codes between two vectors is more efficient due to the order of magnitude change of targets (vector-to-centroid distance calculation). As Fig.\ref{fig:pq-reid} shown, feature vectors in the sub-branch SCR enjoys the same division process of the sub-branch Euclidean. We split the input feature vectors into $M$ sub-spaces and quantize each of them separately. Therefore, each sub-space of the batch image feature vectors are quantized in $C$ centers by k-means algorithm~\cite{1641018}. 
Fig.~\ref{fig:prin} illustrates the detailed process of quantization in SCR. For a given input image $x_i$ encoded as a 2048-d feature vector, the first step is to divide $x_i$ into four sub-spaces yielding a set of sub-features $x_i^j$, where $j$ is the sub-space index. Therefore, each sub-space has a dimension of 512. Within each subspace, image instances are quantized into $C\!=\!256$ centroids. After this, each sub-feature can be represented as an indication of its cluster belonging, which we term the quantization index. The quantization index $C_{x_i^m}^m$ is equivalent to the cluster indicator,  except that it is in a binary-encoded form with the length of 8 bits ($\log_2^{256}$). Concatenating all sub-feature vectors restores a 32 bits whole image representation.

The similarity of image pair in SCR branch is calculated as following:

\begin{equation}
d\left ( x_{i},x_{j} \right )=\sum_{m=1}^{M}d\left ( C_{x_{i}^{m}}^{m},C_{x_{j}^{m}}^{m} \right )
\label{eq:subd}
\end{equation}

where $x_{i}^{m}$ and $x_{j}^{m}$ represent the $m$-th sub-space of $x_{i}$ and $x_{j}$ separately. $C_{x_{i}^{m}}^{m}$ and $C_{x_{j}^{m}}^{m}$ represent the centers these two sub-spaces belong to in $m$-th sub-space. Therefore, the image pair similarity is equal to the summation of all sub-spaces' corresponding centers similarities, which is an approximate distance. The corresponding center represents that the sub-space belongs to.

\begin{figure*}[ht]
\centering
\includegraphics[height=3.8cm,width=9cm]{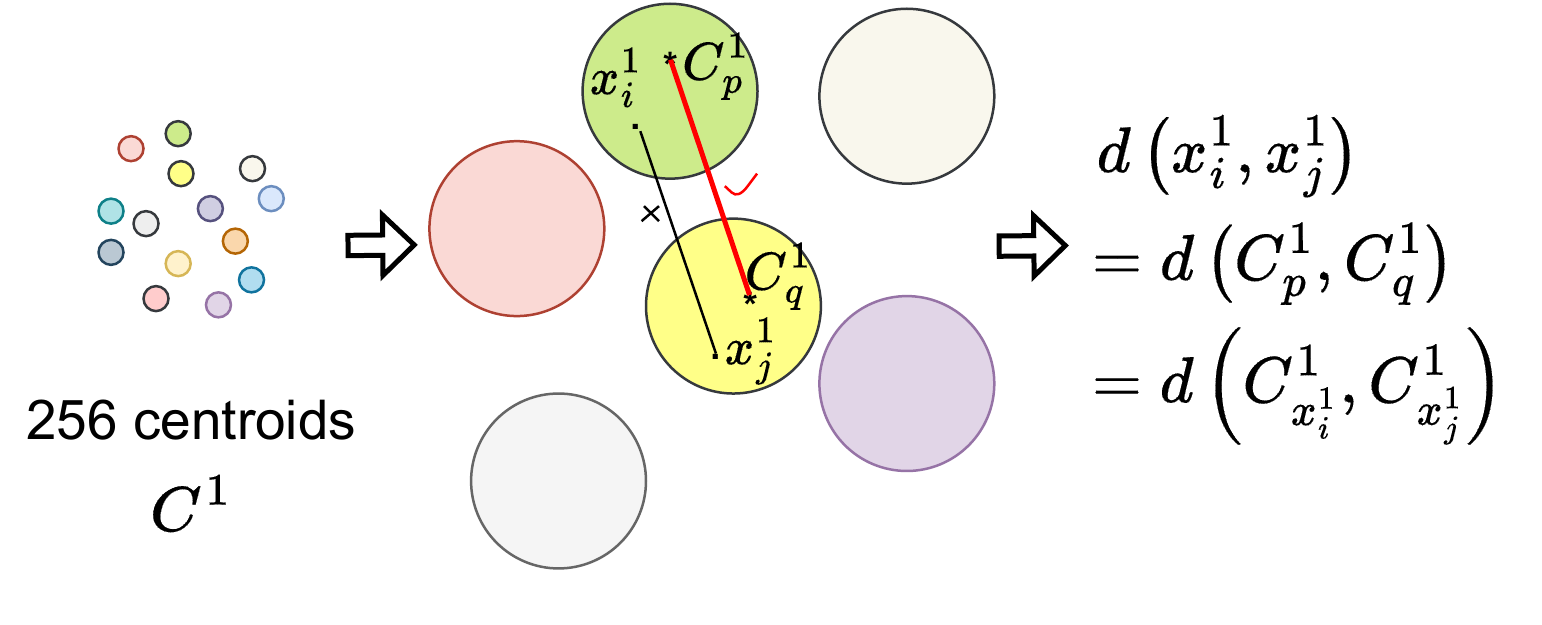}
\caption{Distance calculation based on their quantization indexes. }
\label{fig:calcul}

\end{figure*}

We calculate the image-level similarity between two images as their cumulative distance over all pairs of subspace centroids where sub-features belong. Obtaining the quantized indexes for each image as described above helps us identify corresponding clusters. This distance is simple and efficient as only cluster centroids are used for calculation. Fig.~\ref{fig:calcul} shows a distance calculation example between two quantized feature vectors. In the SCR method, we stored all pre-calculated pairwise distances between centroids in a look-up table to boost the speed further.

Sub-space Consistency Regularization (SCR) algorithm also adopts an offline calculation (pre-calculation before storing) of centroids distances and stores them in a look-up-table ($V_{SCR}$). Based on the optimized principle, SCR vividly avoids slow multiplication operations and leaves only fast summation operations. Thus, distances computation speed is significantly increased compared with the original Euclidean distance.
Besides, due to the dense feature space (centroids are real-value features vector), the accuracy is much higher than binary codes.

Even though, the SCR distance is still in the type of float since the centroids are real-value feature vectors.
This factor prevents the application of the simple counting-sort algorithm.

In online stage, since the Euclidean distance is pre-calculated, the complex multiplication operations is avoided while only simple summation operation is needed. Consequently, the approximate distance computation is very fast.

\textbf{Consistency Regularization}
This section introduces how the proposed consistency regularization loss work. As we mentioned, the float type of SCR distance would prevent the application of counting-sort algorithm.
In order to compare the performance of using integer SCR distance (IntSCR) and SCR distance (SCR), we compare both distances with Consistency Regularization loss separately. Method by adopting SCR distance utilizes quick-sort algorithm to sort, another one by adopting integer SCR distance utilizes counting-sort algorithm.
Here, we choose the MSE loss as the consistency regularization loss to minimize losses and optimize the semantic features training, which is based on SCR distances (from $V_{SCR}$) and shown below:
\begin{equation}
\ell_{cr}=\frac{1}{N^{2}}\sum_{i,j=1}^{N}\left ( \sum_{m=1}^{M}\left \|   V_{SCR}\left ( C_{x_{i}^{m}}^{m}, C_{x_{j}^{m}}^{m}\right )-V_{R}\left (x_{i}^{m},x_{j}^{m}  \right ) \right \|^{2} \right )
\label{eq:pq-loss}
\end{equation}

The consistency regularization loss enhances the quantization influence, which aims to reduce the differences between features generated from CNN (image pair Euclidean distances) and the quantization process (centroid SCR distances), to produce robust and precise cluster centroids.

Combined with the holistic branch, the final loss for our proposed SCR is written as:
\begin{equation}
\ell=\ell_{ce}+\ell_{tri}+\alpha\ell_{cr},
\label{eq:final-loss}
\end{equation}
where $\alpha$ is a hyper-parameter to control the weight of consistency regularization loss.

\textbf{Centroid Update} An iterative framework is proposed to optimize the centroid results and the semantic feature training. In each iteration, the initial cluster centers are the last iteration center results. A parameter of iteration frequency $T$ is used to control this progress. We found empirically that a delayed update of centroids can help stabilize the learning process and achieve better performance.
\subsection{IntSCR for Faster Inference}
\label{subsec:counting-sort}

In branch SCR, we replace the images pair distances with their respective distances of centers. Therefore, a pre-calculated offline look-up-table of centroids Euclidean distances per sub-space is constructed, named $V_{SCR}$, which size is $M \times C \times C$ (as shown in Fig. ~\ref{fig:pq-reid}).
$V_{SCR}$ can achieve a quick image pair distance searching because each feature vector consists of their sub-spaces center indices. The final distance is the summation of each sub-space center distance.
Even though the SCR distance look-up-table speeds distance computation stage up. In ranking stage, it still takes quick-sort algorithm with complexity $O(NlogN)$, where $N$ is the gallery size, because it is still based on \textsl{float} values.
Counting-sort is a ranking algorithm with linear complexity, \textit{i.e.} $O(N)$. However, it requires the array to be ranked in type of \textsl{int}.
Therefore, in this section, we further transfer SCR distance look-up-table to be \textsl{int} ($V_{IntSCR}$ in Fig.~\ref{fig:pq-reid}) and optimize it by a novel iterative framework together with a consistency regularization loss.

\begin{figure*}[ht]
\centerline{\includegraphics[height=4cm,width=8cm]{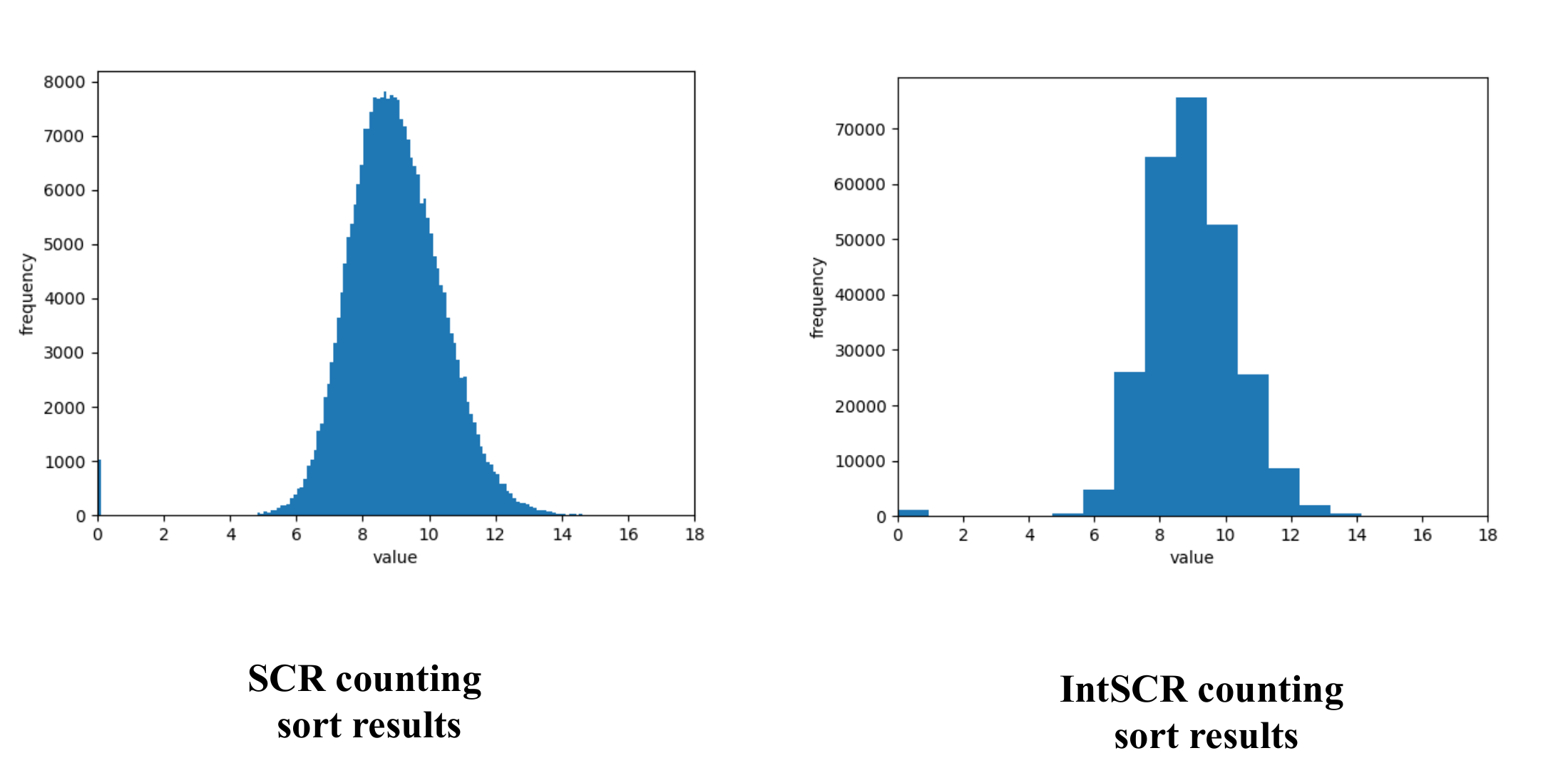}}
\caption{Counting-sort results of float type of SCR and integer type of SCR, where the x-axis represents the distance values and the y-axis represents the counting numbers.}
\label{fig:counting}
\end{figure*}

Fig.~\ref{fig:counting} represents the distribution of distances for float type of SCR distances and integer type of SCR distances. 
As we can see, the counted number of a float type is decuple of integer type. Therefore, the ranking speed of using integer type of SCR distances is much more faster than that of float type, which is verified in Fig.~\ref{fig:speed}.
For the ranking accuracy, integer type of SCR distances cannot significantly drop because the learned centroids are precise and have a distinct disaffinity between each centroid. 
Therefore, the integer type centroids differences are large enough to result in a small influence comparing with the float type of centroid differences in terms of the lost float information. Fig.~\ref{fig:comparison} supports this principle.

For the subspace consistency loss, we can slightly modify Eq.~\eqref{eq:pq-loss} to adapt for $V_{IntSCR}$ by replacing $V_{SCR}$ with $V_{intSCR}$:
\begin{equation}
\ell_{cr}^{'}=\frac{1}{N^{2}}\sum_{i,j=1}^{N}\left ( \sum_{m=1}^{M}\left \|   V_{IntSCR}\left ( C_{x_{i}^{m}}^{m}, C_{x_{j}^{m}}^{m}\right )-V_{R}\left (x_{i}^{m},x_{j}^{m}  \right ) \right \|^{2} \right )
\label{eq:intpq-loss}
\end{equation}

The final loss is then written as below:

\begin{equation}
\ell=\ell_{ce}+\ell_{tri}+\alpha\ell_{cr}^{'}
\label{eq:intfinal-loss}
\end{equation}

\begin{algorithm}
\caption{SCR}
\label{alg:Framwork}
\begin{algorithmic}[1]
\Require 
training data $X=\left \{ x_{i} \right \}_{i=1}^{N}$,
CNN model $\phi \left ( \cdot;\theta \right )$,
iteration frequency $T$,
number of sub-spaces $M$,
number of cluster centers $C$,
weight of consistency regularization loss $\alpha$.
\Ensure optimised CNN model $\hat{\phi} (\cdot ;\hat{\theta }) $
\For{$t$ in range(epoches=120)} 
    \State \textbf{Base CNN Model(Sec.~\ref{subsec:cnn}):}
    \State generate 2048-d feature vector, CNN model learning by cross-entropy and triplet losses.
    \If{$t\quad \text{mod} \quad T=0$}
        \State initialing the cluster center values and repeat the quantization per sub-space.
    \EndIf
    \State \textbf{Sub-space Consistency Regularization (Sec.~\ref{subsec:defin}):}
    \State divide feature vector into multiple sub-spaces and construct $V_{R}$ based on Euclidean distance.
    \State quantize each of sub-space and construct $V_{SCR}$ based on the learned centroids to speed up distance computation.
    \State apply consistency regularization loss to enhance the quantization influence and produce precise cluster centroids.
    \State \textbf{IntSCR for Faster Inference}
    \State transfer $V_{SCR}$ to $V_{IntSCR}$ and apply counting-sort algorithm to speed up rank stage.
    \State train $\phi \left ( \cdot ;\theta^t \right )$ with $X$ and $Y$ with loss in Eq.~\ref{eq:intfinal-loss};
    \State update $Y^t$ with updated clustering centroids in Eq.~\ref{eq:intpq-loss};
\EndFor
\end{algorithmic}
\label{al:training}
\end{algorithm}

\subsection{Learning Scheme}

The detailed training algorithm is presented in Algorithm~\ref{al:training}.
SCR algorithm has an iterative process to update the clustering centers, generating accurate centroids controlled by a weight parameter $T$. This repetition is implemented per $T$ epochs. The centroid results of the current iteration are the initial value of the next clustering.
A precise $V_{SCR}$ is produced to afford a quick distance researching in the distance computing stage. 
Counting-sort algorithm applied for $V_{IntSCR}$ has a significant improvement of speeding up the ranking stage. 
Furthermore, consistency regularization loss enhances the CNN model to maintain the accuracy of ReID. In conclusion, the proposed SCR achieves high speed for evaluating while maintaining the ReID accuracy.

\section{Experiments}
\label{sec:exp}

\subsection{Datasets}
We evaluate the proposed SCR on two representative re-identification datasets.
\textbf{Market-1501}~\cite{7410490} has 32,688 images of 1501 identities which are captured by 6 out-door cameras. The training set consists of 12,936 images of 751 identities. The testing set consists of 19,732 images of 750 identities.
{\bf DukeMTMC-reID}~\cite{zheng2017unlabeled} has 36,411 images of 1,404 identities which are captured by 8 cameras. The training set consists of 16,522 images of 702 identities. The query set has 2,228 images of the other 702 identities. The gallery set has 17,661 images. 

\subsection{Training and Evaluation Protocols}
In the training stage, the iteration frequency of clustering is controlled by a hyper-parameter $T$, and is set to 10 via cross-validation. 
After finishing the training stage, the clustering results of training set will be used for constructing the $V_{IntSCR}$.
Following existing works~\cite{Luo_2019_CVPR_Workshops,wang2020faster}, mean average precision (mAP) and Rank-1 for precision are chosen as the evaluation metrics. We also compared the speed and accuracy of different feature types (real-value feature vector, Hashing binary, and SCR vector) via their corresponding algorithms.

\subsection{Implementation Details}
Following the existing methods \cite{Luo_2019_CVPR_Workshops,wang2020faster}, we set the batch size to 64, input image size is set to $256 \times 128$, training epochs are 120, initial learning rate is $3.5\times 10^{-4}$ and decreased to its $0.1\times$ after 40 epochs and $0.01\times$ after 70 epochs, 10 epochs warm-up is used.
The weight of consistency regularization loss $\alpha$ is 0.01, and the number of cluster centers $C$ is 256. The number of sub-spaces $M$ is adjustable, resulting in a different vector length, which can guarantee a fair comparison with Hashing and real-value algorithms. $C=256=2^{8}$, therefore bits$=8 \times M$. 

\subsection{Parameters Study}
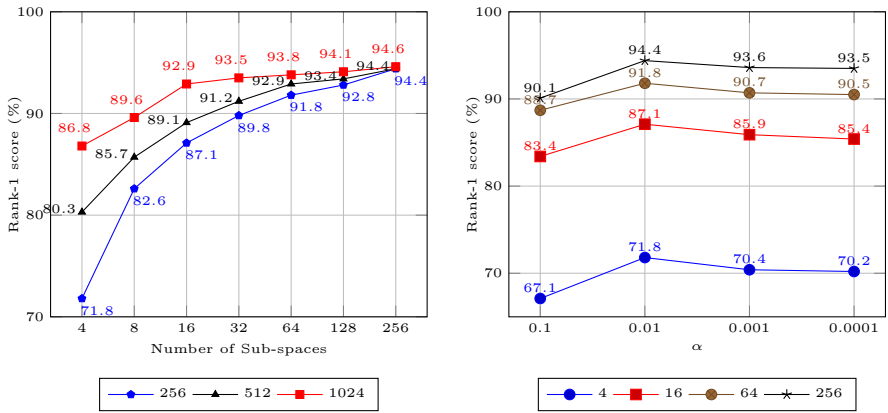
\begin{figure}[!h]
	\centering
	\subfigure[Influence of centers $C$ and subspaces $M$ (where $\alpha$ is 0.01)]{\label{fig:center}
        \begin{tikzpicture}
        	\centering
            \begin{axis}[
            width=0.55\textwidth,
            xmajorgrids,
            ymajorgrids,
            symbolic x coords={4,8,16,32,64,128,256},
            xtick=data,
            nodes near coords,
            node font=\tiny,
            ylabel={Rank-1 score (\%)},
            xlabel={Number of Sub-spaces},
            xlabel near ticks,
            ymin=70, ymax=100,
            y label style={at={(axis description cs:-0.06,.5)},anchor=south},
            legend style={at={(0.5,-0.2)},anchor=north, legend columns=-1},]
            \addplot [color=blue, mark=pentagon*, mark size=1.5pt, every node near coord/.append style={yshift=-0.3cm,xshift=0.2cm}] table [x=x, y=256, col sep=comma] {subspace.dat};
            \addplot [color=black, mark=triangle*,mark size=1.5pt, every node near coord/.append style={yshift=-0.1cm,xshift=-0.3cm}] table [x=x, y=512, col sep=comma] {subspace.dat};
            \addplot[color=red, mark=square*, mark size=1.5pt, every node near coord/.append style={yshift=0.1cm,xshift=-0.1cm}] table [x=x, y=1024, col sep=comma] {subspace.dat};
            \legend{256, 512,1024}
            \end{axis}
        \end{tikzpicture}
	}
	\subfigure[Influence of $\alpha$ (where $C$ is 256)]{ \label{fig:alpha}
        \begin{tikzpicture}
        \centering
            \begin{axis}[
            width=0.55\textwidth,
            symbolic x coords={0.1,0.01,0.001,0.0001},
            xtick=data,
            nodes near coords,
            node font=\tiny,
            xmajorgrids,
            ymajorgrids,
            ymin=65, ymax=100,
            ylabel={Rank-1 score (\%)},
            xlabel near ticks,
            y label style={at={(axis description cs:-0.06,.5)},anchor=south},
            xlabel={$\alpha$},
            legend style={at={(0.5,-0.2)},anchor=north, legend columns=-1},]
            \addplot table [x=x, y=4, col sep=comma] {alpha.dat};
            \addplot table [x=x, y=16, col sep=comma] {alpha.dat};
            \addplot table [x=x, y=64, col sep=comma] {alpha.dat};
            \addplot table [x=x, y=256, col sep=comma] {alpha.dat};
            \legend{4,16,64,256}
            \end{axis}
        \end{tikzpicture}
    }
    
	\caption{Analysis of different parameters setting on Market-1501 with IntSCR distances, including number of centers, subspaces and weight of consistency regularization loss.}
	\label{fig:parameter}
\end{figure}

We vary the weight of our proposed consistency regularization loss, the number of sub-spaces $M$ and clusters $C$, and report the overall performance in Fig.~\ref{fig:parameter}. 

Fig. ~\ref{fig:center} illustrates the different influences by different settings of subspace and centroid numbers. The line in three colours represents corresponding centroid number and the x-axis represents the number of sub-spaces, which can represent its corresponding code length. 
As we can see from Fig.~\ref{fig:center}, larger numbers of cluster centroids can generate high accuracy, especially on low dimensions, but would also result in a high computation consumption. When the number of sub-spaces becomes larger, the rank-1 score gap between different centroids is decreasing. Therefore, we choose the number of centers is 256 in our later experiments.

In Fig. ~\ref{fig:alpha}, we compare different weightings of consistency regularization loss ($\alpha$) with a fixed number of centroids ($C=256$). Different colours represent the different number of sub-spaces.
As we can see from Fig.~\ref{fig:alpha}, the results of all different code length illustrate that $\alpha$ is 0.01 can obtain the best performance when using IntSCR distances.
\begin{table}[h]
\centering
\caption{Comparisons different iteration frequency $T$.}\label{tab:t}
\begin{tabular}{@{\extracolsep{\fill}}lcccc@{\extracolsep{\fill}}}
\toprule%
\multirow{2}{*}[-4pt]{Code Length} & \multicolumn{4}{c}{$T$} \\ \cmidrule(l){2-5} 
 & 1 & 5 & 10 & 20 \\
\midrule
32 & 69.1 & 70.7 & 71.8 & 71.5\\ 
128 & 85.3 & 86.0 & 87.1 & 87.0 \\  
512 & 90.1 & 90.5 & 91.8 & 91.6\\
2048 & 91.2 & 92.9 & 94.4 & 94.1 \\
\botrule
\end{tabular}
\end{table}

In Table.~\ref{tab:t}, we compare different settings of iteration frequency $T$, which represents the times for re-initializing the cluster center values.
It can be inferred that a larger $T$ can help generate better performance when $T <10$, while when $T>10$, the performance drops. Therefore, we choose $T=10$ as our iteration frequency.

\begin{sidewaystable}
\sidewaystablefn%
\begin{center}
\begin{minipage}{\textheight}
\caption{Comparisons with the state-of-the-art non-hashing ReID methods on Market-1501 and DukeMTMC-reID.}\label{tab:nonhash}
\begin{tabular*}{\textheight}{@{\extracolsep{\fill}}lcccccccc@{\extracolsep{\fill}}}
\toprule%
& \multicolumn{2}{@{}c@{}}{Code} & \multicolumn{3}{@{}c@{}}{Market-1501} & \multicolumn{3}{@{}c@{}}{DukeMTMC-reID} \\\cmidrule{2-3}\cmidrule{4-6}\cmidrule{7-9}%
Methods & Type & Length & R1(\%) & mAP(\%) & Q.Time(s) &  R1(\%) & mAP(\%) & Q.Time(s) \\
\midrule
\makecell{PSE}~\cite{Sarfraz2018APE} & R & 1536 & 78.7 & 56.0 & - & - & - & - \\
\makecell{PN-GAN}~\cite{qian2018pose} & R & 1024 & 89.4 & 72.6 & - & 73.6 & 53.2 & - \\
\makecell{IDE}~\cite{zheng2016person} & R & 2048 & 88.1 & 72.8 & - & 69.4 & 55.4 & -\\
\makecell{Camstyle}~\cite{zhong2018camera} & R & 2048 & 88.1 & 68.7 & - & 75.3 & 53.5 & - \\
\makecell{BoT}~\cite{Luo_2019_CVPR_Workshops} & R & 2048 & 94.1 & \underline{85.7} & $2.200$ & \underline{86.4} & \underline{76.4} & $2.000$\\
\midrule
\makecell{SPReID}~\cite{8578215} & R & 10,240 & 92.5 & 81.3 & - & 84.4 & 71.0 & - \\
\makecell{VPM}~\cite{8953249} & R & 14,336 & 93.0 & 80.8 & - & 83.6 & 72.6 & - \\
\makecell{PCB}~\cite{pcb2018} & R & 12,288 & 93.8 & 81.6 & $6.900$ & 83.3 &69.2 & $6.300$\\
\midrule
\makecell{IntSCR}(ours)  & B & 1024 & 92.8 & 83.3 & $0.205$ & 85.7 & 72.3 & $0.211$ \\
\makecell{IntSCR}(ours)  & B & 2048 & \underline{94.4} & 84.3 & $0.558$ & 85.9 & 72.5 & $0.456$ \\
\botrule
\end{tabular*}
\footnotetext{ `Q.Time' represents researching time per query image.
'R' represents the real-value feature which demonstrates that longer features produce higher accuracy but with a slower speed.
`IntSCR', integer type of our proposed SCR distance, shows its balancing ability between keep comparable accuracy with non-hashing methods and quicker speed.}
\end{minipage}
\end{center}
\end{sidewaystable}

\begin{sidewaystable}
\sidewaystablefn%
\begin{center}
\begin{minipage}{\textheight}
\caption{Comparisons with the state-of-the-art hashing ReID methods on Market-1501 and DukeMTMC-reID. 'B' represents the binary code.}\label{tab:hash}
\begin{tabular*}{\textheight}{@{\extracolsep{\fill}}lcccccccc@{\extracolsep{\fill}}}
\toprule%
& \multicolumn{2}{@{}c@{}}{Code} & \multicolumn{3}{@{}c@{}}{Market-1501} & \multicolumn{3}{@{}c@{}}{DukeMTMC-reID} \\\cmidrule{2-3}\cmidrule{4-6}\cmidrule{7-9}%
Methods & Type & Length & R1(\%) & mAP(\%) & Q.Time(s) &  R1(\%) & mAP(\%) & Q.Time(s) \\
\midrule
\makecell{DRSCH}~\cite{7185403} & B & 512 & 17.1 & 11.5 & - & 19.3 & 13.6 & -\\
\makecell{HashNet}~\cite{8237860} & B & 512 & 29.2 & 19.1 & - & 40.8 & 28.6 & -\\
\makecell{CSBT}~\cite{8100049} & B & 512 & 42.9 & 20.3 & - & 47.2 & 33.1 & - \\
\makecell{TDDH}~\cite{8876675} & B & 512 & 74.5 & 25.9 & $0.010$ & 65.7 & 27.8 & -\\
\makecell{ABC}~\cite{8784980} & B & 512 & 69.4 & 48.5 & $0.098$ & 69.9 & 52.6 & $0.075$\\
\makecell{ABC}~\cite{8784980} & B & 2048 & 81.4 & 64.7 & $0.280$ & 82.5 & 61.2 & $0.200$\\
\makecell{CtF}~\cite{wang2020faster} & B & 32 & 60.0 & 37.7 & $0.034$ & 49.5 & 28.7 & $0.023$\\
\makecell{CtF}~\cite{wang2020faster} & B & 128 & 88.9 & 71.0 & $0.042$ & 78.6 & 59.4 & $0.032$\\
\makecell{CtF}~\cite{wang2020faster} & B & 512 & 92.8 & 82.2 & $0.098$ & 85.4 & 71.6 & $0.075$\\
\makecell{CtF}~\cite{wang2020faster} & B & 2048 & 93.7 & \underline{85.4} & $0.280$ & \underline{87.7} & \underline{75.7} & $0.200$\\
\midrule
\makecell{IntSCR}(ours) & B & 32 & 71.8 & 63.3 & $0.015$ & 65.0 & 63.4 & $0.009$ \\
\makecell{IntSCR}(ours)  & B & 128 & 87.1 & 80.7 & $0.040$ & 81.4 & 67.0 & $0.031$ \\
\makecell{IntSCR}(ours) & B & 512 & 91.8 & 82.5 & $0.099$ & 84.8 & 71.6 & $0.094$ \\
\makecell{IntSCR}(ours)  & B & 1024 & 92.8 & 83.3 & $0.205$ & 85.7 & 72.3 & $0.211$ \\
\makecell{IntSCR}(ours)  & B & 2048 & \underline{94.4} & 84.3 & $0.558$ & 85.9 & 72.5 & $0.456$ \\

\botrule
\end{tabular*}
\end{minipage}
\end{center}
\end{sidewaystable}

\subsection{Comparison with the State-of-the-Arts}
\label{subsec:comp}

We compare our proposed method IntSCR (integer type of SCR distances) with three real-value feature methods (BoT~\cite{Luo_2019_CVPR_Workshops}, PN-GAN~\cite{qian2018pose}, PCB~\cite{pcb2018}) and two binary code methods (CtF~\cite{wang2020faster}, TDDH~\cite{8876675}). Experimental results are shown in Table~\ref{tab:nonhash} and Table.~\ref{tab:hash}.
Generally speaking, binary code is very fast but accuracy is relatively lower especially under short code, real-value features get high accuracy even under short feature but speed down a lot.
Our proposed IntSCR significantly outperforms binary codes meanwhile keeps very fast speed, beats real-value features in speed. In conclusion, the proposed IntSCR achieves a better balance between speed and accuracy than both real-value features and binary codes.

\textbf{Comparison with real-value features.}
Generally, a longer feature contributes to a higher accuracy but usually with a slower speed. BoT~\cite{Luo_2019_CVPR_Workshops} achieves the results of R1 $94.1\%$ and mAP $85.7\%$ but with a lower speed $2.2s$ on Market-1501, and results of R1 $86.4\%$ and mAP $76.4\%$ but with a lower speed $2s$ on DukeMTMC-reID. These BoT results outperform IntSCR (2048-d) in R1 $0.5\%$, mAP $3.9\%$ on DukeMTMC-reID and mAP $1.4\%$ on Market-1501. But R1 of BoT on Market-1501 is lower than that of IntSCR by $0.3\%$ and speeds of BoT are $3.9\times/4.3\times$ slower than that of IntSCR ($0.56s/0.46s$) on Market-1501 and DukeMTMC-reID, respectively.

Table~\ref{tab:nonhash} illustrates that IntSCR (1024-d) outperforms PN-GAN~\cite{qian2018pose} in both R1 by $3.4\%$ and mAP by $8.9\%$ on Market-1501 and R1 by $12.1\%$ and mAP by $19.1\%$ on DukeMTMC-reID. We compare with PCB~\cite{pcb2018} which has 12,288-dimensional features. Table~\ref{tab:nonhash} shows that IntSCR (2048-d) outperforms PCB in R1 by $0.6\%/2.6\%$ and mAP by $2.7\%/3.3\%$, speed of IntSCR are $0.08\times/0.07\times$ quicker than that of PCB on Market-1501 and DukeMTMC-reID, respectively.

\textbf{Comparison with binary codes.} 
In Hashing ReID methods, CtF~\cite{wang2020faster} with 2048-dimensional binary codes achieves the results of R1 $93.7\%$, mAP $85.4\%$ and considerable speed $2.8\times10^{-1}$ on Market-1501, results of R1 $87.7\%$, mAP $75.7\%$ and considerable speed $2\times10^{-1}$ on DukeMTMC-reID. Table.~\ref{tab:hash} shows that CtF outperforms IntSCR (2048-d) in R1 by $1.8\%$, mAP by $3.2\%$, speed $0.44\times$ on DukeMTMC-reID and mAP by $1.1\%$, speed $0.5\times$ on Market-1501. But its R1 on Market-1501 is lower than IntSCR by $0.7\%$. On low-dimensional binary codes (32 bits), Table.~\ref{tab:hash} shows that IntSCR (32-d) outperforms CtF on both R1 $11.8\%$ and mAP $25.6\%$ and speed $0.44\times$ on Market-1501 and R1 $15.5\%$ and mAP $34.7\%$ and speed $0.39\times$ on DukeMTMC-reID. IntSCR (512-d) also outperforms TDDH~\cite{8876675} which has 512-dimensional binary codes in R1 by $17.3\%$ and mAP by $56.6\%$ on Market-1501 and R1 by $19.1\%$ and mAP by $43.8\%$ on DukeMTMC-reID. However, IntSCR has a slower speed than TDDH which is $9.9\times$ on Market-1501.
In conclusion, in terms of image searching speed, IntSCR is quicker than real-value methods and low-dimensional binary codes but slower than Hashing. For the accuracy of low dimensions features, IntSCR outperforms both the real-value method (PN-GAN) and Hashing (CtF, TDDH) by a large margin.
\begin{figure}[!h]
	\centering
	\subfigure[Rank-1]{\label{fig:rank}
        \begin{tikzpicture}
            \begin{axis}[width=0.55\textwidth, height=0.5\textwidth,
            ybar=1pt, 
            bar width=3pt,
            symbolic x coords={32,64,128,256,512,1024,2048},
            xtick=data,
            nodes near coords,
            node font=\tiny,
            nodes near coords style={anchor=west,rotate=90},
            ylabel={Rank-1 score (\%)},
            xlabel={Code Length},
            xtick pos=left,
            ytick pos=left,
            xlabel near ticks,
            y label style={at={(axis description cs:-.05,.5)},anchor=south},
            ymin=50, ymax=103,
            legend pos=south east,
            legend cell align={left},]

            \addplot table [x=x, y=euclidean, col sep=comma] {rank.dat};
            \addplot table [x=x, y=hamming, col sep=comma] {rank.dat};
            \addplot table [x=x, y=scr, col sep=comma] {rank.dat};
            \addplot table [x=x, y=intscr, col sep=comma] {rank.dat};

            \legend{Euclidean, Hamming, SCR, IntSCR};
            \end{axis}
        \end{tikzpicture}
    }
	\subfigure[mAP]{\label{fig:map}
        \hspace{-0.5em}
        \begin{tikzpicture}
            \begin{axis}[
            width=0.55\textwidth,
            height=0.5\textwidth,
            ybar=1pt, 
            bar width=3pt,
            symbolic x coords={32,64,128,256,512,1024,2048},
            xtick=data,
            xtick pos=left,
            ytick pos=left,
            nodes near coords,
            node font=\tiny,
            nodes near coords style={anchor=west,rotate=90},
            ylabel={mAP score (\%)},
            xlabel={Code Length},
            xlabel near ticks,
            y label style={at={(axis description cs:-0.05,.5)},anchor=south},
            ymin=35, ymax=95,
            legend pos=south east,
            legend cell align={left},]
            \addplot table [x=x, y=euclidean, col sep=comma] {map.dat};
            \addplot table [x=x, y=hamming, col sep=comma] {map.dat};
            \addplot table [x=x, y=scr, col sep=comma] {map.dat};
            \addplot table [x=x, y=intscr, col sep=comma] {map.dat};
            \legend{Euclidean,Hamming , SCR, IntSCR};
            \end{axis}
        \end{tikzpicture}
    }
	\caption{Performance comparisons of distance types on Market-1501, including Hamming, Euclidean, SCR and IntSCR. }
	\label{fig:comparison}
\end{figure}
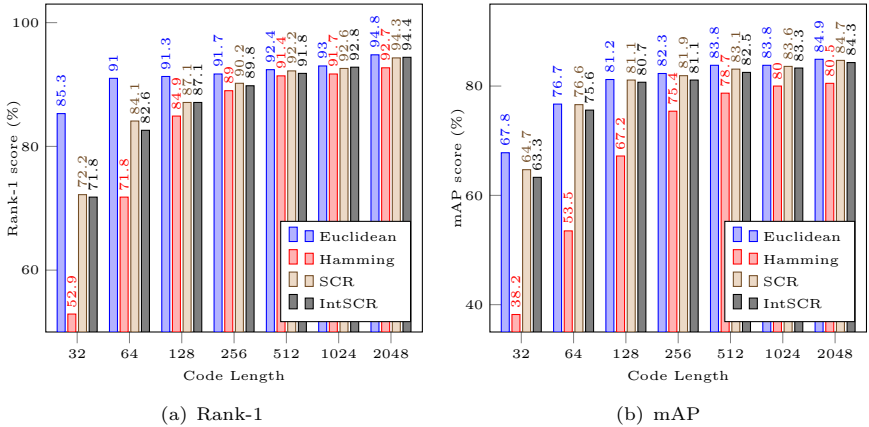

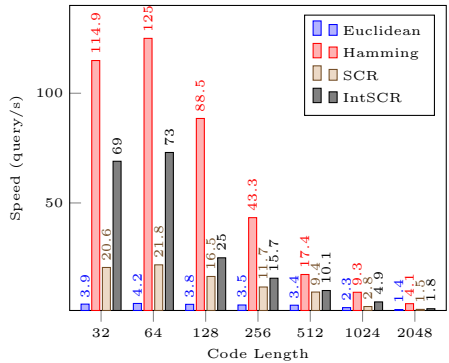
\begin{figure}[!h]
	\centering
        \hspace{-2em}\begin{tikzpicture}
            \begin{axis}[
            ybar=1pt, 
            width=0.55\textwidth,
            bar width=3pt,
            symbolic x coords={32,64,128,256,512,1024,2048},
            xtick=data,
            nodes near coords,
            node font=\tiny,
            nodes near coords style={anchor=west,rotate=90},
            ylabel={Speed (query/s)},
            xlabel={Code Length},
            xlabel near ticks,
            ylabel near ticks,
            xtick pos=left,
            ytick pos=left,
            ymin=1, ymax=140,
            legend pos= north east,
            legend cell align={left},]

            \addplot table [x=x, y=euclidean, col sep=comma] {speed.dat};
            \addplot table [x=x, y=hamming, col sep=comma] {speed.dat};
            \addplot table [x=x, y=msq, col sep=comma] {speed.dat};
            \addplot table [x=x, y=intmsq, col sep=comma] {speed.dat};
            
            \legend{Euclidean,Hamming,SCR,IntSCR};
            
            \end{axis}
        \end{tikzpicture}
	\caption{Speed comparisons of distance types on Market-1501, including Hamming, Euclidean, SCR and IntSCR. IntSCR gets a better balance between accuracy and speed.}
	\label{fig:speed}
\end{figure}

Fig.~\ref{fig:comparison} and Fig.~\ref{fig:speed} show the compare four different distance methods, \ie, {\bf Hashing, real-value, SCR and IntSCR}). Fig.~\ref{fig:speed} shows that SCR and IntSCR can achieve significant speed improvements compared to real-value counterparts. The performances are comparable to Hashing, especially on low dimensions. 
The speed of IntSCR is $17\times$ faster than Euclidean with the code length being 32-d. The reason is that the computation complexity of Euclidean is higher than IntSCR, and also, quick-sort requires more time than counting-sort. With the increase in code length, all methods slow down in speed. That is because the distance calculation cost increases with the feature dimension. Fig.~\ref{fig:comparison} shows that SCR and IntSCR can achieve performance close to real-value and have an impressive advantage over Hashing on low dimensions. The Rank-1 accuracy of IntSCR is $1.4\times$ that of Hamming distance, while mAP is $1.7\times$ higher with a 32-d code length. This improvement is not surprising as the IntSCR distance is based on real-value features.
In contrast, Hamming distance operates on binary codes. 
As code length increases, all methods' performance rises because more discriminative features are used as input.
Such comparison demonstrates the effectiveness of our proposed IntSCR in addressing the Fast ReID problem. We conclude that the Euclidean distance method achieves the best performance with the slowest speed. While Hamming distance method obtains the best speed but the worst performance across all code lengths. Our proposed IntSCR distance method achieves a good trade-off between performance and speed.

\begin{figure*}[t]
\centerline{\includegraphics[height=6cm,width=12cm]{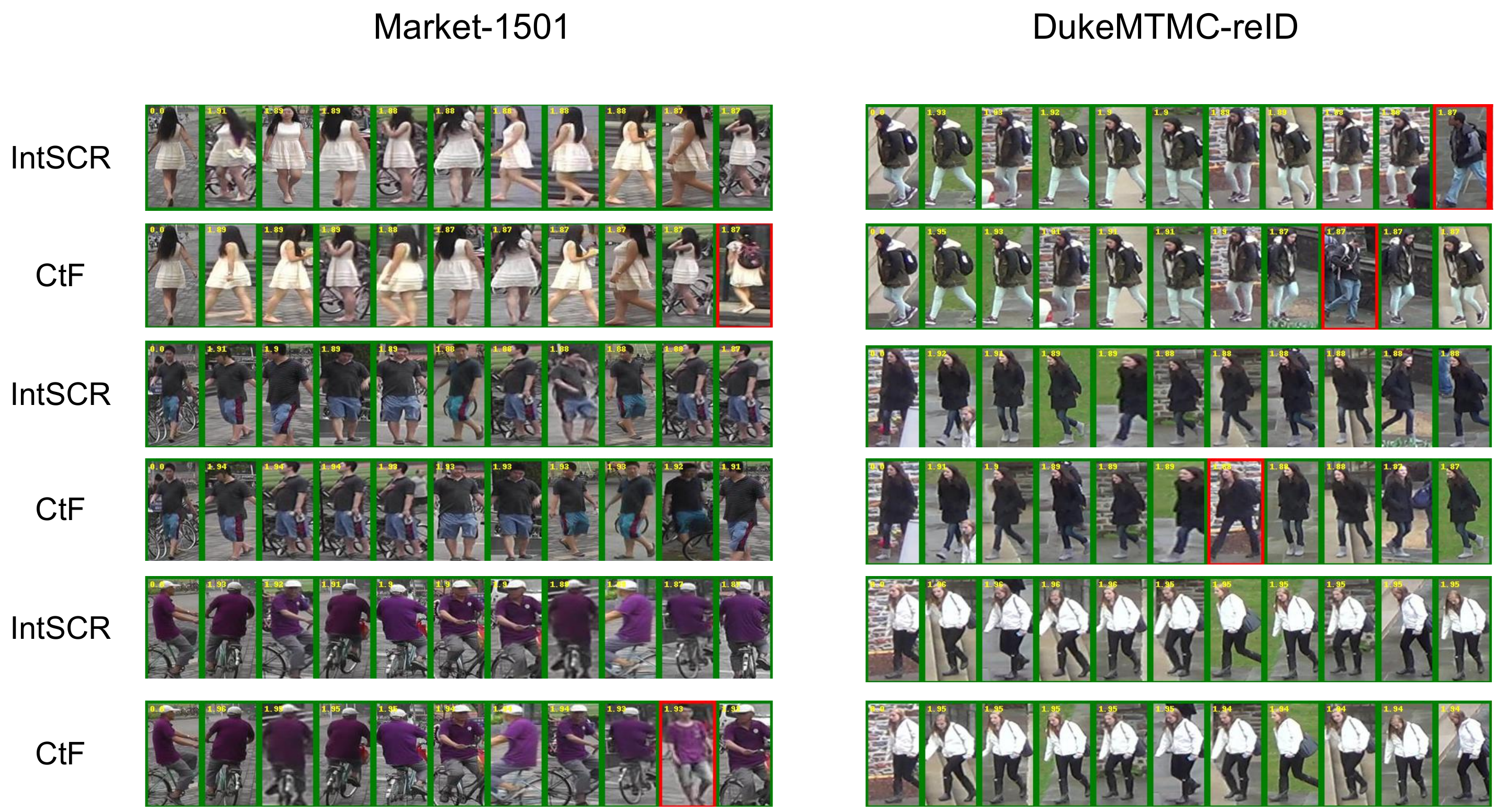}}

\caption{Ranking lists comparisons of two methods on Market-1501 and DukeMTMC-reID datasets.}
\label{fig:meth-data}
\end{figure*}

Fig.~\ref{fig:meth-data} visualizes the outputs from different methods on two benchmarks. We compared CtF~\cite{wang2020faster} to our proposed IntSCR with the same experimental setting with 2048-d feature inputs. We plot top-10 ranking results for three randomly sampled query images. It can be seen that both methods obtain excellent ranking lists with only extremely few mismatches. Thus, our proposed IntSCR demonstrates a competitive retrieving ability compared to the state-of-the-art CtF~\cite{wang2020faster}.

\subsection{Evaluation on Large-Scale Dataset}
\label{subsec:largedata}

\begin{figure*}[t]
\centerline{\includegraphics[height=6cm,width=15cm]{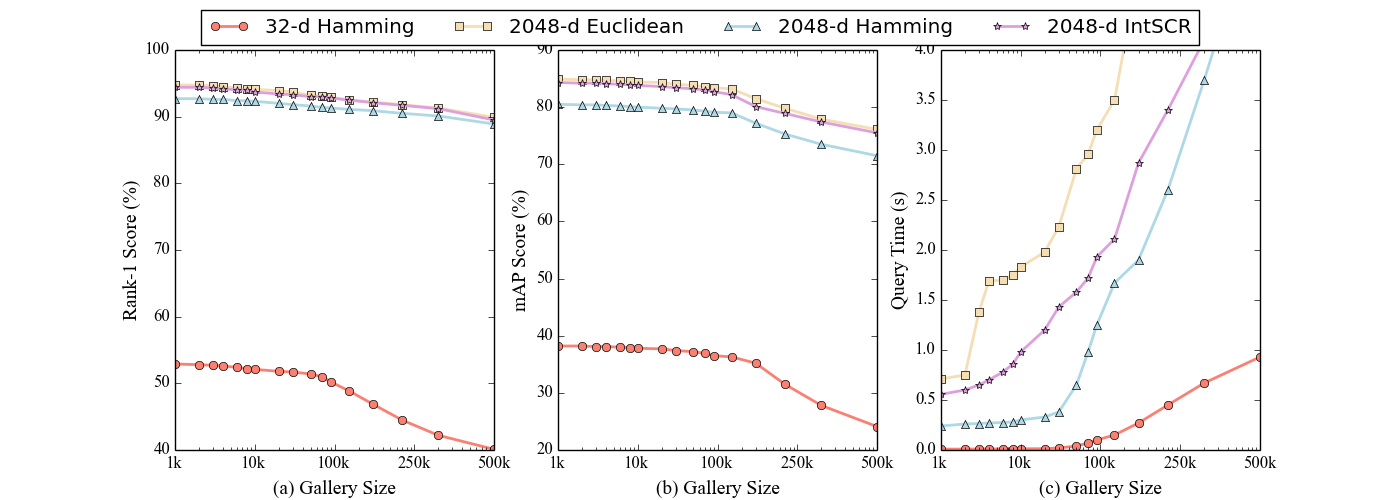}}

\caption{Compared different methods on large-scale dataset Market-1501+500k.}
\label{fig:largedata}
\end{figure*}

The size of the gallery matters when evaluating the accuracy and speed of a faster ReID approach. We further apply our approach and other algorithms on a larger-scale ReID dataset, \ie, Market-1501+500K~\cite{7410490}.   Market-1501+500K extends the widely-used Market-1501~\cite{7410490} dataset by providing 500k more bounding box annotations. The experimental results are reported in Fig.~\ref{fig:largedata}.

As the gallery size increases, Rank-1 and mAP accuracies of all methods experienced different degrees of decrease, and the speed of image retrieval also slows down. This is expected as more gallery images make ReID more challenging. Besides, an increase in gallery images accounts for more distance computing and sorting time. 
Similar to what we observed on a smaller dataset, the 2048-d Euclidean distance method obtains the best accuracy but has the worst searching time. 
Two Hamming distance methods take less time than other methods because counting sort and hamming distance calculation of binary codes are faster to compute. At the same time, 2048-d binary code requires more time than 32-d binary code simply because 32-d is easier. Our proposed method IntSCR still manages to obtain a comparable accuracy to the Euclidean distance algorithm and a comparable searching time to hamming distance algorithm. That demonstrates that our proposed method remarkably balances accuracy and time on a larger-scale dataset.

\subsection{Backbones}
\label{subsec:model_analysis}

We further study the generalizability of our proposed method by incorporating it with multiple backbone architectures. These backbones include VGG-19~\cite{vgg19}, ResNet-18~\cite{7780459}, ResNet-50~\cite{7780459} and ResNet-101~\cite{7780459}.
We append our proposed IntSCR on top of different backbones and report the results with 32-d feature inputs in Table~\ref{tab:models}.
As is reported, ResNet-101 acquires the best accuracy but the worst speed. VGG-19 produces the worst accuracy, and ResNet-18 obtains the worst speed. In conclusion, models that are deeper in layers generally yield higher performance but slower speed. That is because more layers allow for more fine-tuning but spend more time on computing. Compared to other backbone options, ResNet-50 achieves a better balance between accuracy and speed.

\begin{table}[h]
\centering
\caption{Comparisons of different models with IntSCR algorithm on Market-1501.}\label{tab:models}
\begin{tabular}{@{\extracolsep{\fill}}cccc@{\extracolsep{\fill}}}
\toprule%
Model Type & Rank-1(\%) & mAP(\%) & Speed(query/s) \\
\midrule
VGG-19 & 56.2 & 51.1 & 73\\ 
ResNet-18 & 61.1 & 61.6 & 76\\  
ResNet-50 & 71.8 & 63.3 & 69\\
ResNet-101 & 72.1 & 65.7 & 59\\
\bottomrule
\end{tabular}
\end{table}

\section{Conclusions}
\label{sec:con}
This work proposes a novel model named SCR for improving ReID matching speed whilst maintaining accuracy, especially on low-dimensional feature representations, to solve the disadvantage of Hashing and real-value feature distance metrics. We explore SCR on ReID to generate an effective distance table for image searching speed. Clustering per sub-space is implemented to achieve precise distances for keeping high accuracy. The effectiveness of SCR is evidenced through extensive comparative evaluations.
Our proposed method strikes a decent balance between algorithm accuracy and efficiency. However, as we observe, there is still a performance gap compared to real-value feature methods and a speed gap compared to Hashing methods. In the future, we will continue to investigate how to improve the ReID performance while achieving faster speed.

\bibliographystyle{bst-sn-basic}
\bibliography{sn-bibliography}
\end{document}